\def\year{2020}\relax
\documentclass[letterpaper]{article} 
\usepackage{aaai20}  
\usepackage{times}  
\usepackage{helvet} 
\usepackage{courier}  
\usepackage[hyphens]{url}  
\usepackage{graphicx} 
\usepackage{graphicx}  
\title{Towards Building a Multilingual Sememe Knowledge Base:\\
Predicting Sememes for BabelNet Synsets}

\author{
Fanchao Qi$^{1}\thanks{Indicates equal contribution}$,
Liang Chang$^{2*}\thanks{Work done during internship at Tsinghua University}$, 
Maosong Sun$^{13}\thanks{Corresponding author}$,
Sicong Ouyang$^{2\dag}$, 
Zhiyuan Liu$^{1}$ 
\\ 
$^{1}$Department of Computer Science and Technology, Tsinghua University \\
Institute for Artificial Intelligence, Tsinghua University \\
Beijing National Research Center for Information Science and Technology \\
$^{2}$Beijing University of Posts and Telecommunications \\
$^{3}$Jiangsu Collaborative Innovation Center for Language Ability, Jiangsu Normal University\\
{qfc17@mails.tsinghua.edu.cn, changliang@bupt.edu.cn}\\
{ \{sms, liuzy\}@tsinghua.edu.cn, scouyang4354@gmail.com}
}

\usepackage{epsfig}
\usepackage{amsmath,amssymb}
\usepackage{multirow,booktabs, hhline}
\usepackage{placeins}

\newcommand{\citet}[1]{\citeauthor{#1} \shortcite{#1}}
\newcommand{\citep}{\cite}

\begin{document}
\maketitle
\begin{abstract}

A sememe is defined as the minimum semantic unit of human languages. 
Sememe knowledge bases (KBs), which contain words annotated with sememes, have been successfully applied to many NLP tasks. 
However, existing sememe KBs are built on only a few languages, which hinders their widespread utilization.
To address the issue, we propose to build a unified sememe KB for multiple languages based on BabelNet, a multilingual encyclopedic dictionary. 
We first build a dataset serving as the seed of the multilingual sememe KB. 
It manually annotates sememes for over $15$ thousand synsets (the entries of BabelNet).
Then, we present a novel task of automatic sememe prediction for synsets, aiming to expand the seed dataset into a usable KB.
We also propose two simple and effective models, which exploit different information of synsets.
Finally, we conduct quantitative and qualitative analyses to explore important factors and difficulties in the task.
All the source code and data of this work can be obtained on \url{https://github.com/thunlp/BabelNet-Sememe-Prediction}.
\end{abstract}

\section{Introduction}

A word is the smallest element of human languages that can stand by itself, but not the smallest indivisible semantic unit. 
In fact, the meaning of a word can be divided into smaller components. 
For example, one of the meanings of ``man'' can be represented as the composition of the meanings of ``human'', ``male'' and ``adult''. 
In linguistics, a \emph{sememe} \citep{bloomfield1926set} is defined as the minimum semantic unit of human languages. 
Some linguists believe that meanings of all the words in any language can be decomposed of a limited set of predefined sememes, which is related to the idea of universal semantic primitives \citep{wierzbicka1996semantics}.

\begin{figure}[t]
	\includegraphics[width=.95\columnwidth]{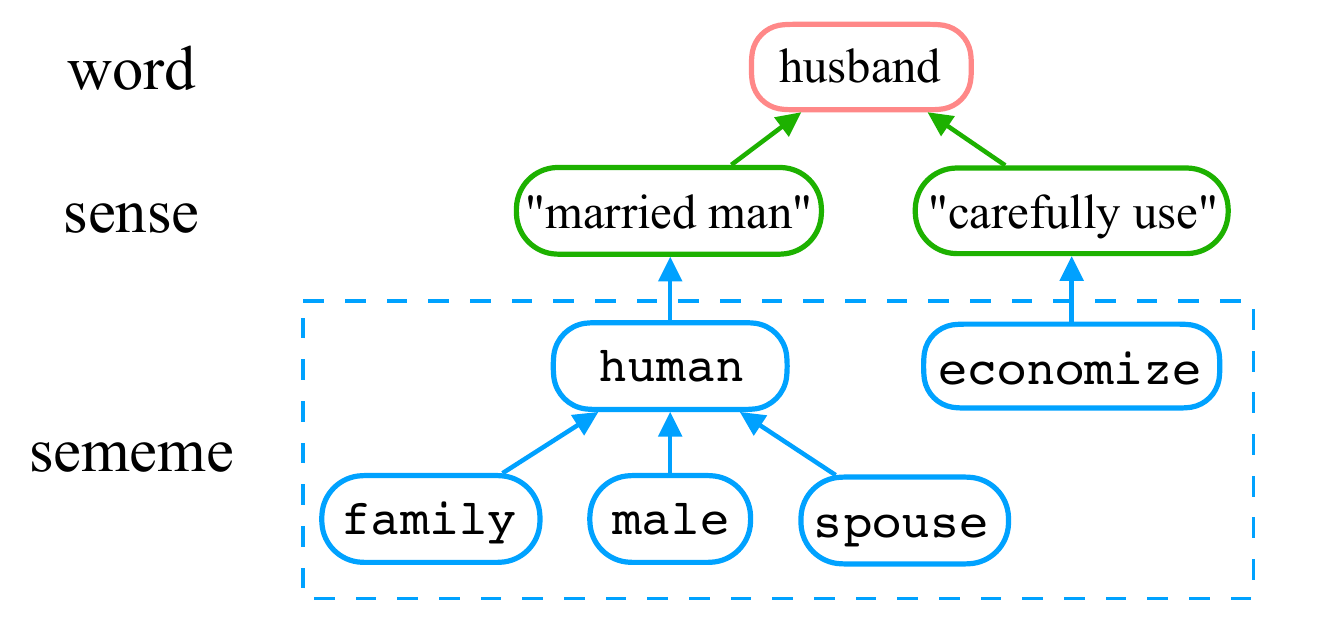}	
    \caption{Sememe annotation of the word ``husband'' in HowNet.} 
	\label{fig:hownet} 
\end{figure}
Sememes are implicit in words. 
To utilize them in practical applications, people manually annotate words with predefined sememes to construct sememe knowledge bases (KBs). 
HowNet \citep{dong2003hownet} is the most famous one, 
which uses about $2,000$ language-independent sememes to annotate senses of over 100 thousand Chinese and English words. 
Figure \ref{fig:hownet} illustrates an example of how words are annotated with sememes in HowNet.

Different from most linguistic KBs like WordNet \citep{miller1995wordnet}, which explain meanings of words by word-level relations, sememe KBs like HowNet provide intensional definitions for words using infra-word sememes. 
Sememe KBs have two unique strengths.
The first one is their sememe-to-word semantic compositionality, which endows them with special suitability for integration into neural networks \citep{gu2018language,qi2019modeling}.
The second one is their nature that limited sememes can represent unlimited meanings, which makes sememes very useful in low data regimes, e.g., improving embeddings of low-frequency words \citep{sun2016embedding,niu2017improved}. 
In fact, sememe KBs have been proven beneficial to various NLP tasks such as word sense disambiguation \citep{duan2007word} and sentiment analysis \citep{fu2013multi}. 

Most languages have no sememe KBs, which prevents NLP applications of these languages benefiting from sememe knowledge. 
However, building a sememe KB for a new language from scratch is time-consuming and labor-intensive --- the construction of HowNet takes several linguistic experts more than two decades. 
To tackle this challenge, \citet{Qi2018Cross} present the task of cross-lingual lexical sememe prediction (CLSP), aiming to
facilitate the construction of a new language's sememe KB by predicting sememes for words in that language. 
However, CLSP can predict sememes for only one language at a time, which means repetitive efforts, including manual correctness checking conducted by native speakers, are required when constructing sememe KBs for multiple languages.


\begin{figure}[t]
	\includegraphics[width=.95\columnwidth]{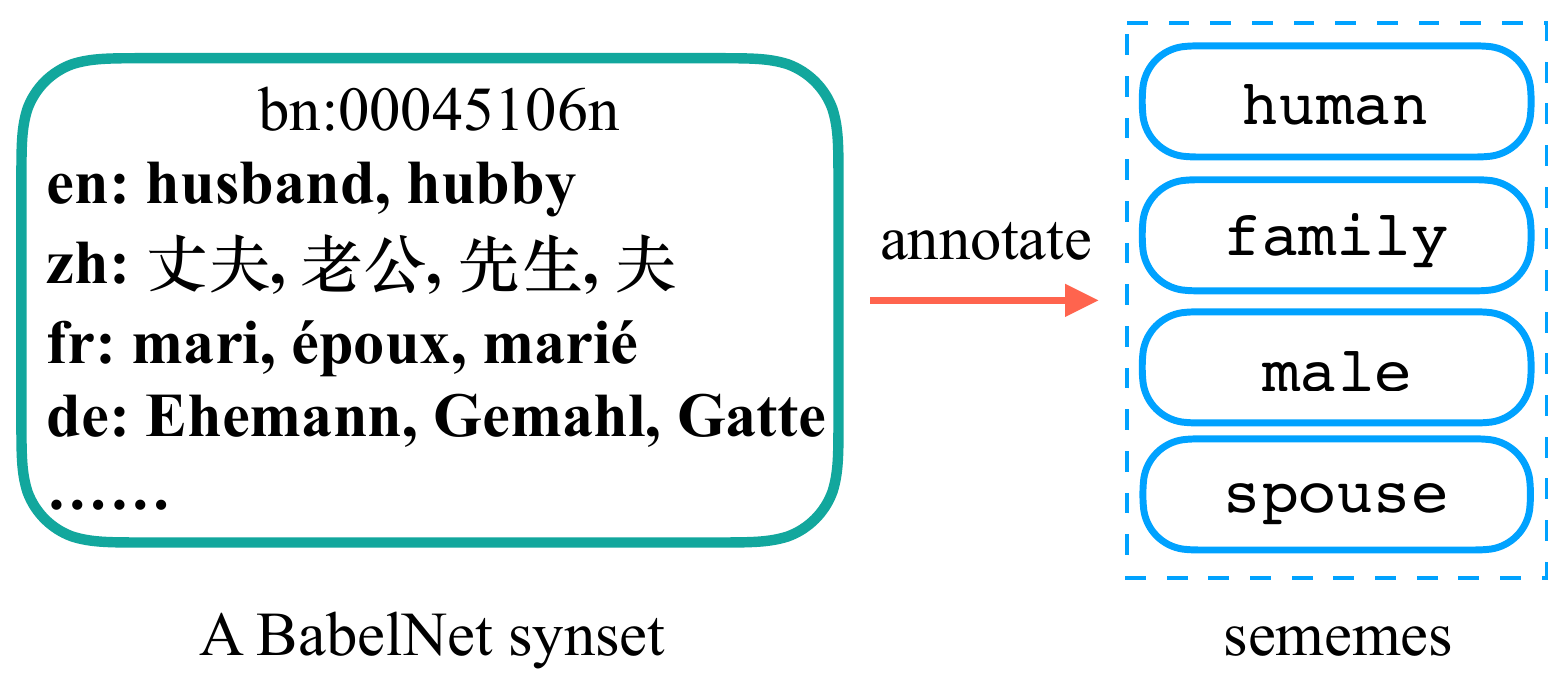}	
    \caption{Annotating sememes for the BabelNet synset whose ID is \textit{bn:00045106n}. The synset comprises words in different languages (multilingual synonyms) having the same meaning ``the man a woman is married to'', and they share the four sememes on the right part.}
	\label{fig:BabelNet}
\end{figure}

To solve this problem,
we propose 
to build a unified sememe KB for multiple languages, namely a multilingual sememe KB, based on BabelNet \citep{navigli2012babelnet}, which is a more economical and efficient way to transfer sememe knowledge to other languages.
BabelNet is a multilingual encyclopedic dictionary and comprises over 15 million entries called \textit{BabelNet synsets}.
Each BabelNet synset contains words in multiple languages with the same meaning (multilingual synonyms), and they should have the same sememe annotation. 
Therefore, building a multilingual sememe KB by annotating sememes for BabelNet synsets can actually provide sememe annotation for words in multiple languages simultaneously (Figure \ref{fig:BabelNet} shows an example). 

To advance the creation of a multilingual sememe KB, 
we build a seed dataset named \textbf{BabelSememe}, which contains about 15 thousand BabelNet synsets manually annotated with sememes. 
Also, we present a novel task of automatic sememe prediction for BabelNet synsets (SPBS), aiming to gradually expand the seed dataset into a usable multilingual sememe KB.  
In addition, we exploratively put forward two simple and effective models for SPBS, 
which utilize different information incorporated in BabelNet synsets.
The first model exploits semantic information and recommends similar sememes to semantically close BabelNet synsets, 
while the second 
uses relational information and directly predicts relations between sememes and BabelNet synsets. 
In experiments, we evaluate sememe prediction performance of the two models on BabelSememe, finding they achieve satisfying results. 
Moreover, the ensemble of the two models yield obvious performance enhancement. 
Finally, we conduct detailed quantitative and qualitative analyses to explore the factors influencing sememe prediction results, aiming to point out the characteristics and difficulties of the SPBS task.

In conclusion, our contributions are threefold: 
(1) first proposing to construct a multilingual sememe KB based on BabelNet and presenting a novel task SPBS; 
(2) building the BabelSememe dataset containing over 15 thousand BabelNet synsets manually annotated with sememes; 
and (3) proposing two simple and effective models and conducting detailed analyses of factors in SPBS.

\section{Dataset and Task Formalization}

\subsection{BabelSememe Dataset}
\label{sec:babelsememe}

We build the BabelSememe dataset, which is expected to be the seed of a multilingual sememe KB and expanded steadily by automatic sememe prediction together with examination of humans.
Sememe annotation in HowNet embodies hierarchical structures of sememes, as shown in Figure \ref{fig:hownet}.
Nevertheless, considering the structures of sememes are seldom used in existing applications \citep{niu2017improved,qi2019modeling} and it is very hard for ordinary people to make structured sememe annotation, we ignore them in BabelSememe. 
Thus, each BabelNet synset in BabelSememe is annotated with a set of sememes (as shown in Figure \ref{fig:BabelNet}). 
In the following part, we elaborate how we build the dataset and provide its statistics.

\paragraph{Selecting Target Synsets}
We first select 20 thousand synsets as target synsets.
Each of them
includes English and Chinese synonyms annotated with sememes in HowNet,\footnote{We use OpenHowNet \citep{qi2019openhownet}, the open-source data accessing API of HowNet to obtain the sememes of a word.} so that we can generate candidate sememes for them.

\paragraph{Generating Candidate Sememes} 
We generate candidate sememes for each target synset using the sememes annotated to its synonyms.
Some sememes of the synonyms in a target synset should be annotated to the target synset. 
For example, the word ``husband'' is annotated with five sememes in HowNet (Figure \ref{fig:hownet}) in total, and the four sememes annotated to the sense ``married man'' should be annotated to the BabelNet synset \textit{bn:00045106n} (Figure \ref{fig:BabelNet}). 
Thus, we group the sememes of all the synonyms in a target synset together to form the candidate sememe set of the target synset.

\paragraph{Annotating Appropriate Sememes} 
We ask more than 100 annotation participants to select appropriate sememes from corresponding candidate sememe set for each target synset.
All the participants have a good grasp of both Chinese and English.
We show them Chinese and English synonyms as well as definitions of each synset, making sure its meaning is fully understood. 
When annotation is finished, each target synset has been annotated by at least three participants. 
We remove the synsets whose inter-annotator agreement (IAA) is poor, where we use Krippendorff's alpha coefficient \citep{krippendorff2013content} to measure IAA.
Finally, $15,756$ BabelNet synsets are retained, which are annotated with $43,154$ sememes selected from $104,938$ candidate sememes, and their average Krippendorff's alpha coefficient is $\textbf{0.702}$. 


\begin{table}[t!]
\centering
\resizebox{.95\columnwidth}{!}{
    \begin{tabular}{cccccc}
        \toprule
        POS Tag & noun & verb & adj & adv & total \\
        \midrule
        $\#$synset  & 10,417 & 2,290 & 2,507 &  542 & 15,756 \\
        \midrule
        average $\#$sememe & 2.95& 2.49 & 2.29 & 1.78 & 2.74 \\
        \bottomrule
    \end{tabular}
}
\caption{Statistics of BabelNet synsets with different POS tags in BabelSememe.}
\label{tab:stats}
\end{table}

\begin{figure}[!t]
    \centering
    \includegraphics[width=.93\columnwidth]{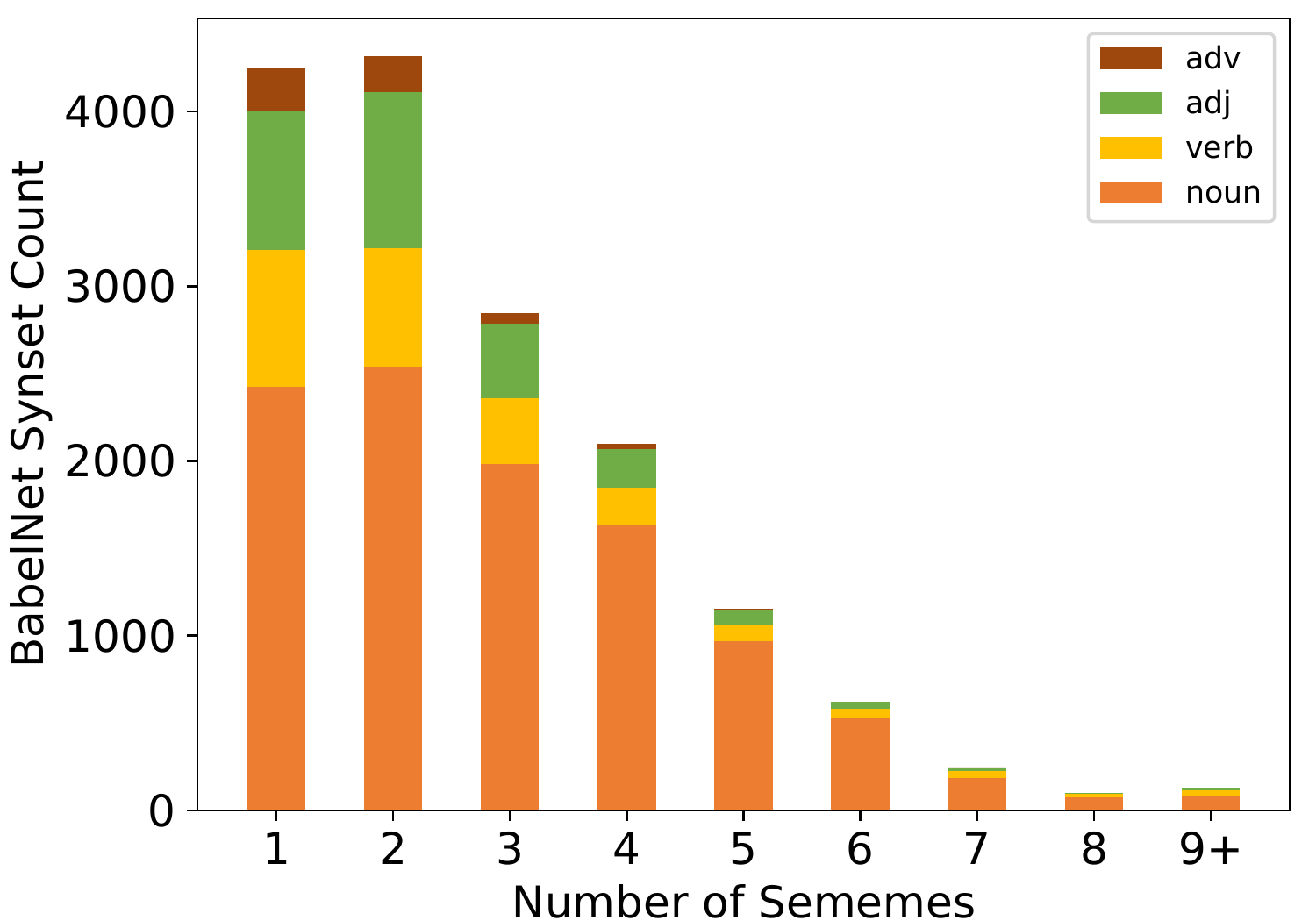}
    \caption{Distribution of BabelNet synsets over sememe numbers in BabelSememe.}
    \label{fig:babel-stat1}
\end{figure}

\paragraph{Dataset Statistics}
Detailed statistics of BabelNet synsets with different POS tags in BabelSememe are shown in Table \ref{tab:stats}. 
In addition, we show the distribution of BabelNet synsets over sememe numbers in Figure \ref{fig:babel-stat1}.

\subsection{SPBS Task Formalization}
\label{sec:task}
SPBS is aimed at predicting appropriate sememes for unannotated BabelNet synsets. 
It can be modeled as a multi-label classification problem, where sememes are regarded as labels to be attached to BabelNet synsets. 
Formally, we define $B$ as the set of all the BabelNet synsets, and $S$ as the set of all the sememes. 
For a given target BabelNet synset $b \in B$, we intend to predict its sememe set $S_b=\{s_1,\cdots,s_{|S_b|}\}\subset S$, where
$|S_b|$ is the number of $b$'s sememes.


Previous methods of sememe prediction for \textit{words} usually compute an association score for each sememe and select the sememes with scores higher than a threshold to form the predicted sememe set \citep{xie2017lexical,jin2018incorporating}. 
Following this formulation, we have 
\begin{equation}
    \hat{S}_b=\{s\in S|P(s|b) > \delta\},
    \label{eq:main}
\end{equation}
where $\hat{S}_b$ is the predicted sememe set of $b$, $P(s|b)$ is the association score of sememe $s$, and $\delta$ is the association score threshold. 
To compute the association score, existing methods of sememe prediction for \textit{words} capitalize on semantic similarity between target words and sememe-annotated words or directly model the relations between target words and sememes \citep{xie2017lexical,jin2018incorporating}. 
No matter which way to choose, representations of target words are of vital importance. 
Similarly, representations of BabelNet synsets are crucial to SPBS. 
In following section of Methodology, we make an preliminary attempt to utilize two kinds of representations of BabelNet synsets in the SPBS task.

\section{Methodology}
\label{sec:method}



As mentioned in task formalization, learning representations of BabelNet synsets\footnote{From now on, we use ``synset'' as the abbreviation of ``BabelNet synset'' for succinctness.} is very important to SPBS.
BabelNet merges various resources such as WordNet \citep{miller1995wordnet} and Wikipedia, which provide abundant information for learning synset representations.
Correspondingly, we summarize two kinds of synset representations according to the information used for representation learning:
(1) \textbf{semantic representation}, which bears the meaning of a synset. 
Much information can be used for learning the semantic representation of a synset, e.g., 
textual definitions from WordNet and related Wikipedia articles;
(2) \textbf{relational representation}, which captures the relations between different synsets. Most relations are the semantic relations transferred from WordNet (e.g., ``antonym'').

Next, we introduce two preliminary models, namely SPBS-SR and SPBS-RR, which utilize semantic and relational representations respectively to predict sememes for synsets.
We also present an ensemble model which combines the two models' prediction results to obtain better performance.



\subsection{SPBS-SR Model}
Inspired by \citet{xie2017lexical}, the idea of SPBS-SR (SPBS with Semantic Representations) is to compute the association score $P(s|b)$ by measuring the similarity between the target synset $b$ and the other synsets annotated with the target sememe $s$. 
In other words, if a synset with known sememe annotation is very similar to the target synset, its sememes should have high association scores. 
In fact, this idea is similar to collaborative filtering \citep{sarwar2001item} in recommendation systems.

Formally, following the notations in task formalization, we can calculate $P(s|b)$ by
\begin{equation}
\begin{aligned}
\label{eq:cr}
    P(s|b) \sim \sum_{b'\in B'}{\boldmath \cos(\mathbf{b},\mathbf{b}') \cdot 
    I_{S_{b'}}(s)
    \cdot c^{r_{b'}}},
\end{aligned}
\end{equation}
where $B'$ is the set of synsets with known sememe annotation, $\mathbf{b}$ and $\mathbf{b}'$ represent the semantic representations of $b$ and $b'$ respectively, $I_{S_{b'}}(s)$ is a indicator function indicating whether $s$ is in $S_{b'}$, $r_{b'}$ is the descending rank of $\cos(\mathbf{b},\mathbf{b}')$ and $c \in (0, 1)$ is a hyper-parameter. Here $c^{r_{b'}}$ is a declined confidence factor used to diminish the influence of irrelevant words on predicting sememes.  The irrelevant words may have totally different sememes which are actually noises.

We choose the embedded vector representations of synsets from NASARI \citep{camacho2016nasari} as required semantic representations.
NASARI utilizes the content of Wikipedia pages linked to synsets to learn vector representations, and only nominal synsets have NASARI representations because non-nominal synsets have no linked Wikipedia pages.
Correspondingly, SPBS-SR can be used in sememe prediction for nominal synsets only.

\subsection{SPBS-RR Model}
\label{sec:SPBS-RR}

\begin{figure}[t]
	\includegraphics[width=.95\columnwidth]{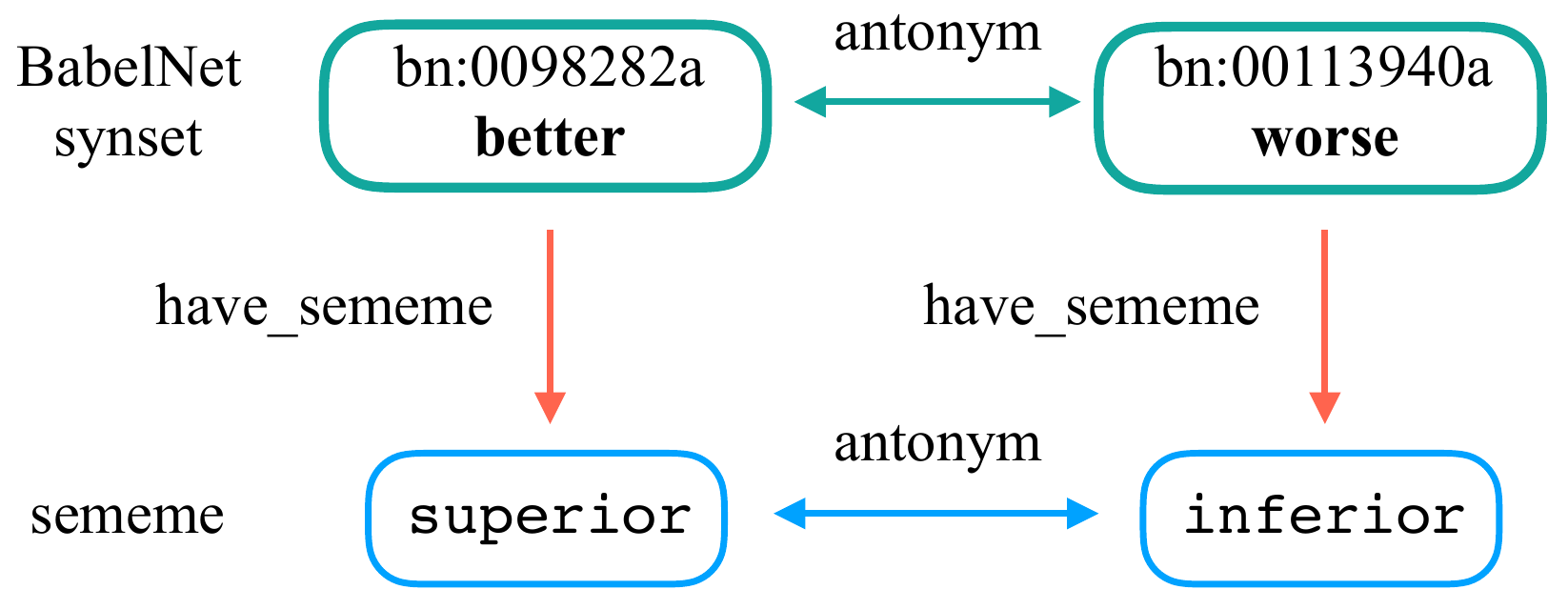}	
	\caption{An example of how relations between BabelNet synsets are consistent with the relations between respective sememes. Notice that we only show the English synonyms in the BabelNet synsets.}
	\label{fig:example} 
\end{figure}

SPBS-RR (SPBS with Relational Representations) aims to use relational representations of target synsets to compute $P(s|b)$.
As mentioned before, there are many relations between synsets, and most of them are semantic relations from WordNet. 
As for sememes, they also have four semantic relations in HowNet, namely ``hypernym'', ``hyponym'', ``antonym'' and ``converse''.
The semantic relation between a pair of synsets should be consistent with the relation between their respective sememes. 
Taking Figure \ref{fig:example} as an example, the synset ``better'' is the antonym of ``worse'', and their respective sememes \texttt{superior} and \texttt{inferior} are also a pair of antonyms.
Naturally, this property can be used to predict sememes when synsets and sememes as well as all their relations are considered together --- if we know ``better''-``worse'' and \texttt{superior}-\texttt{inferior} are both antonym pairs, and ``better'' has the sememe \texttt{superior}, then we should undoubtedly predict \texttt{inferior} for ``worse''.

To this end, we introduce an artificial relation ``have\_sememe'' between a synset and any of its sememes, aiming to build a bridge between synsets and sememes. 
Now by considering all the synset-synset, sememe-sememe and synset-sememe relations, synsets and sememes form a semantic graph, and as a result, knowledge graph embedding methods can be used to learn relational representations of synsets.

Here we borrow the translation idea from the well-established TransE model \citep{Bordes2013}. Formally, for the above-mentioned semantic graph, all its relation triplets form a set $G$.
Each triplet in $G$ can be represented as $(h,r,t)$, where $h,t \in S \cup B$ are nodes and $r\in R_B \cup R_S \cup \{r_h\}$ is a relation. $R_B$ and $R_S$ are the sets of relations between synsets and between sememes respectively, and $r_h$ refers to the ``have\_sememe'' relation. 
Then we can learn representations of both nodes and relations by minimizing:
\begin{equation}
\begin{aligned}
\label{eq:rr-loss1}
\mathcal{L}_{1} = \sum_{(h,r,t)\in G} [\tau + d(\mathbf{h}+\mathbf{r},\mathbf{t})-d(\mathbf{h}+\mathbf{r},\mathbf{t}')]_+,
\end{aligned}
\end{equation}
where $[x]_+ = \max(0,x)$, scalar $\tau$ is a hyper-parameter, $(h,r,t'), t'\in S\cup B$ is a corrupted triplet, boldface symbols represent corresponding vectors, and $d(\mathbf{x},\mathbf{y})$ is $L_2$ distance function:
\begin{equation}
\begin{aligned}
\label{equ:eq3}
d(\mathbf{x},\mathbf{y})=\|\mathbf{x}-\mathbf{y}\|^2.
\end{aligned}
\end{equation}

However, the synset-sememe semantic graph is not exactly the same as general knowledge graphs, because it embraces two kinds of nodes, namely synsets and sememes.
Furthermore, according to the definition of sememes, the meaning of a synset should be equal to the sum of its sememes' meanings.
In other words, there exists a special semantic equivalence relation between a synset and all its sememes. 
We formalize this relation and design a corresponding constraint loss:
\begin{equation}
\begin{aligned}
\mathcal{L}_{2}=\sum_{b\in B}\|\mathbf{b}+\mathbf{r}_s-\sum_{s\in S_b}\mathbf{s} \|^2,
\end{aligned}
\end{equation}
where $r_s$ denotes the semantic equivalence relation.
Therefore, the overall training loss is as follows:
\begin{equation}
\begin{aligned}
\label{eq:rr-loss}
\mathcal{L} = \lambda_1 \mathcal{L}_{1} + \lambda_2 \mathcal{L}_{2},
\end{aligned}
\end{equation}
where $\lambda_1$ and $\lambda_2$ are hyper-parameters controlling relative weights of the two losses.
By optimizing the loss function, we can obtain relational representations of synsets, sememes and relations. 
Sememe prediction for the target synset $b$ can be regarded as finding the tail node of the incomplete triplet $(b,r_h,?)$.
Hence, the association score $P(s|b)$ can be computed by:
\begin{equation}
\begin{aligned}
\label{equ:eq1}
    P(s|b) \sim d(\mathbf{b}+\mathbf{r}_h,\mathbf{s}).
\end{aligned}
\end{equation}

\subsection{Ensemble Model}
SPBS-SR depends on semantic representations of synsets while SPBS-RR utilizes relational representations.
It is obvious that the two models employ different kinds of information and exhibit different features of a synset.
Accordingly, combining the two models together is expected to improve sememe prediction performance. 
Considering sememe association scores yielded by the two models are not comparable, we redefine the sememe prediction score by making use of the reciprocal sememe ranks:
\begin{equation}
\begin{aligned}
\label{eq:ensemble}
P(s|b) = \lambda_c\frac{1}{{\rm rank}^c_s} + \lambda_r\frac{1}{\rm{rank}^r_s},
\end{aligned}
\end{equation}
where ${\rm rank}^c_s$ and ${\rm rank}^r_s$ are descending ranks of sememe $s$ according to the association scores computed by SPBS-SR and SPBS-RR respectively, and $\lambda_c$ and $\lambda_r$ are hyper-parameters controlling relative weights of the two items.

\begin{table*}[!t]
  \centering
    \begin{tabular}{lcccccccccc}
    \toprule
    \multicolumn{1}{c}{\multirow{2}[4]{*}{Model}} & \multicolumn{2}{c}{noun} & \multicolumn{2}{c}{verb} & \multicolumn{2}{c}{adj} & \multicolumn{2}{c}{adv} & \multicolumn{2}{c}{avg.} \\
\cmidrule{2-11}          & MAP   & F1    & MAP   & F1    & MAP   & F1    & MAP   & F1    & MAP   & F1 \\
    \midrule
    LR & 54.4  & 40.0  & ---   & ---   & ---   & ---   & ---   & ---   & ---   & --- \\
    TransE & 60.2  & 46.6  & 33.2  & 24.6  & 31.1  & 24.3  & 30.0  & \bfseries{21.3}  & 51.3  & 39.5 \\
    \hline
    SPBS-SR & 65.0  & 50.0  & ---   & ---   & ---   & ---   & ---   & ---   & ---   & --- \\
    SPBS-RR & 62.5  & 47.9  & \textbf{34.8} & \textbf{25.3}  & \textbf{32.7}  & \textbf{24.5}  & \textbf{30.9}  & 20.0  & 53.3  & 40.5  \\
    \hline
    Ensemble & \bfseries{69.0}  & \bfseries{55.4} & \bfseries{34.8}  & \bfseries{25.3}  & \bfseries{32.7}  & \bfseries{24.5}  & \bfseries{30.9}  & 20.0  & \bfseries{57.6}  & \bfseries{45.6}  \\
    \bottomrule
    \end{tabular}%
\caption{Overall and POS tag-specific SPBS results of all the models on the test set.}
\label{tab:results-all}%
\end{table*}%

\section{Experiments}
In this section, we evaluate our two SPBS models. Furthermore, we conduct both quantitative and qualitative analyses to investigate the factors in SPBS results, aiming to reveal characteristics and difficulties of the SPBS task. 


\subsection{Dataset}

We use BabelSememe as the source of sememe annotation for synsets. 
We extract all the relations between the synsets in BabelSememe from BabelNet.
In addition, there are four semantic relations between sememes, and the ``have\_sememe'' relation between a synset and any of its sememes.
Since our SPBS-RR model is graph-based, following previous knowledge graph embedding work, we filter the low-frequency synsets, sememes, and relations out.\footnote{Although our SPBS-SR model can handle the low-frequency synsets well, we use the same filtered dataset to evaluate both models for fair comparison.}

The final dataset we use contains $15,461$ synsets, $2,106$ sememes, $196$ synset-synset relations, 4 sememe-sememe relations and 1 synset-sememe relation (``have\_sememe'').
And there are $171,147$ triplets in total, including $125,114$ synset-synset, $3,317$ sememe-sememe and $42,716$ synset-sememe triplets. 
We randomly divide the synsets into three subsets in the ratio of 8:1:1.
Since only the tail nodes of synset-sememe triplets need to be predicted in the SPBS-RR model, we select all the synset-sememe triplets comprising the synsets in the two 10\% subsets to form the validation and test sets. 
All the other triplets compose the training set.

Notice that our first model SPBS-SR only works on nominal synsets and needs no synset-synset or sememe-sememe triplets, which means only the synset-sememe triplets comprising nominal synsets in the training set are utilized and only nominal synsets 
have sememe prediction results.

\subsection{Experimental Settings}
\paragraph{Baseline Methods}
SPBS is a brand new task, and there are no previous methods specifically designed for it.
Hence, we simply choose logistic regression (LR) and TransE\footnote{Numerous knowledge graph embedding methods have been proposed recently, but we find that TransE performs substantially better than all other popular models on this task by experiment.} as baseline methods. 
Similar to SPBS-SR, LR also takes NASARI embeddings of synsets (semantic representations) as input and only works on nominal synsets. 
TransE learns relational representations of synsets and differs from SPBS-RR in the constraint of semantic equivalence relation.

\paragraph{Hyper-parameters and Training}
For both SPBS-SR and LR, the dimension of used NASARI synset embeddings is $300$.
For SPBS-SR, $c$ in Equation \eqref{eq:cr} is $0.8$.
For both TransE and SPBS-RR, 
the embedding dimension of synsets, sememes and relations is empirically set to $800$, and the margin $\tau$ in Equation \eqref{eq:rr-loss1} is set to $4$. 
For SPBS-RR, the relative weight $\lambda_1 = 0.95$ and $\lambda_2 = 0.05$.
For the ensemble model, $\lambda_c = 0.45$ and $\lambda_r = 0.55$. 
We adopt SGD as the optimizer whose learning rate is set to $0.01$.
All these hyper-parameters have been tuned to the best on the validation test. 

\paragraph{Evaluation Metrics}
Following previous sememe prediction work, we choose mean average precision (MAP) and the F1 score as evaluation metrics. 
And the sememe association score threshold, 
i.e., $\delta$ in Equation \eqref{eq:main}, is set to 0.32.

\begin{table}[!t]
\centering
\resizebox{.95\columnwidth}{!}{
    \begin{tabular}{ccccc}
    \toprule
    POS Tag   & noun  & verb  & adj   & adv \\
    \midrule
    $\#$synset & 10,360 & 2,240  & 2,419  & 442 \\
    $\#$triplet & 210,127 & 20,657 & 23,490 & 4,952 \\
    avg. $\#$triplet  & 20.28  & 9.22  & 9.71  & 11.20  \\
    \bottomrule
    \end{tabular}
}
\caption{Numbers of POS tag-specific synsets and triplets and their average triplet numbers.}
\label{tab:stat-ex}
\end{table}

\subsection{Overall SPBS Results}

The overall and POS tag-specific SPBS results of our models as well as baseline methods on the test set are shown in Table \ref{tab:results-all}. 
Note that the ensemble model has the same results as SPBS-RR on non-nominal synsets because SPBS-SR works on nominal synsets only.
From the table, we can see that:

(1) SPBS-RR performs markedly better than TransE on synsets with whichever POS tag. This can prove the effectiveness of the constraint of semantic equivalence relation, which we propose specifically for the SPBS task by taking advantage of the nature of sememes. 

(2) On the nominal synsets, SPBS-SR achieves the best performance among all of the four single models, which demonstrates that recommending identical sememes to meaning-similar synsets is effectual. 
In addition, the ensemble model produces substantial performance improvement as compared to its two submodels, which manifests the success of our ensemble strategy. 

(3) SPBS-RR yields much better results on nominal synsets than non-nominal synsets. 
To explore the reason, we count the number of synsets with different POS tags as well as the numbers of triplets comprising POS tag-specific synsets and calculate their average triplet numbers.
The statistics are listed in Table \ref{tab:stat-ex}.
We find that the number of nominal synsets and their average triplet number are significantly bigger than those of the non-nominal synsets. 
Consequently, less relational information of non-nominal synsets is captured, and it is hard to learn good relational representations for them, which explains their bad performance. 

\subsection{SPBS for Nominal Synsets}
According to the statistics in Table \ref{tab:stat-ex}, we speculate those non-nominal synsets have a negative influence on sememe prediction for nominal synsets.
To prove this, we remove all the non-nominal synsets as well as related triplets from the dataset, and then re-evaluate all the models on nominal synsets. 
The results are shown in Table \ref{tab:ex-noun}. 

\begin{table}[!t]
  \centering
    \begin{tabular}{lrr}
    \toprule
    model & \multicolumn{1}{l}{MAP} & \multicolumn{1}{l}{F1} \\
    \midrule
    LR & 59.5 & 45.3 \\
    TransE & 65.2  & 50.4  \\
    SPBS-SR & 65.0  & 50.0  \\
    SPBS-RR & {66.0}  & {51.0}  \\
    Ensemble & \bfseries{70.3}  & \bfseries{56.7}  \\
    \bottomrule
    \end{tabular}%
    \caption{Sememe prediction results of all the models on nominal synsets.}
\label{tab:ex-noun}%
\end{table}%

We observe that both TransE and SPBS-RR receive considerable performance boost in predicting sememes for nominal synsets, and they even produce better results than SPBS-SR. 
In addition, the ensemble model performs correspondingly better. 
These results confirm our previous conjecture, and point out that we should notice the effect of other low-frequency synsets on the target synsets in SPBS.

\begin{figure}[!t]
    \centering
    \includegraphics[width=.92\columnwidth]{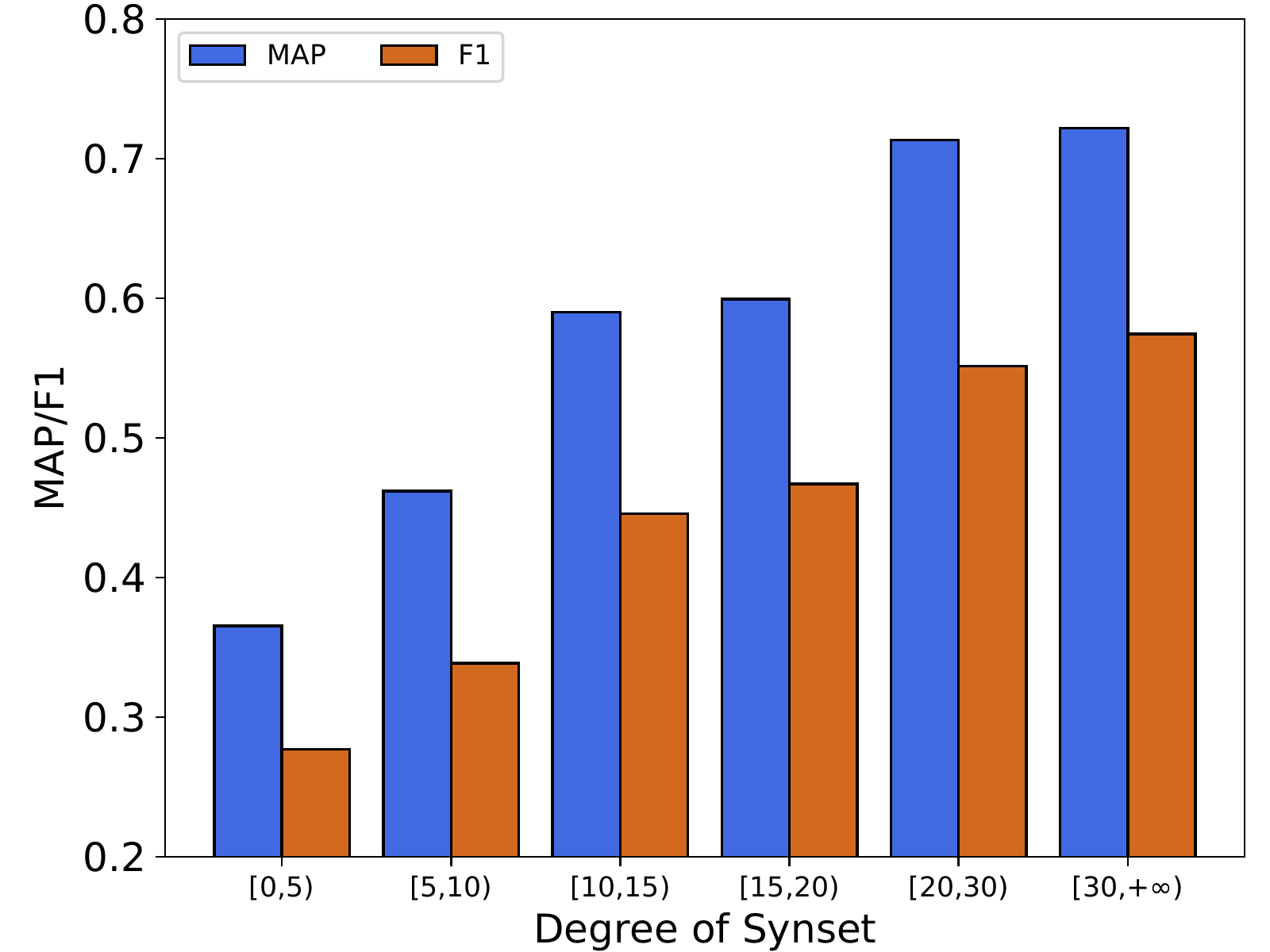}
    \caption{SPBS results of synsets within different degree ranges. The numbers of synsets in the six ranges are 72, 340, 231, 110, 84 and 131 respectively.}
    \label{fig:eff-synset-degree}
\end{figure}

\subsection{Effect of the Synset's Degree}
In this subsection, we investigate the effect of the synset's degree on SPBS results, where the degree of a synset is the number of the triplets comprising the synset. This experiment as well as following ones is conducted on the nominal synsets of the test set using the ensemble model.

Figure \ref{fig:eff-synset-degree} exhibits
the sememe prediction results of synsets within different degree ranges. 
We find that the degree of a synset has great impact on its sememe prediction results. 
It is easier to predict sememes for the synsets with larger degrees because more relation information of these synsets is captured and better relational representations are learned.

\begin{figure}[!t]
    \centering
    \includegraphics[width=.92\columnwidth]{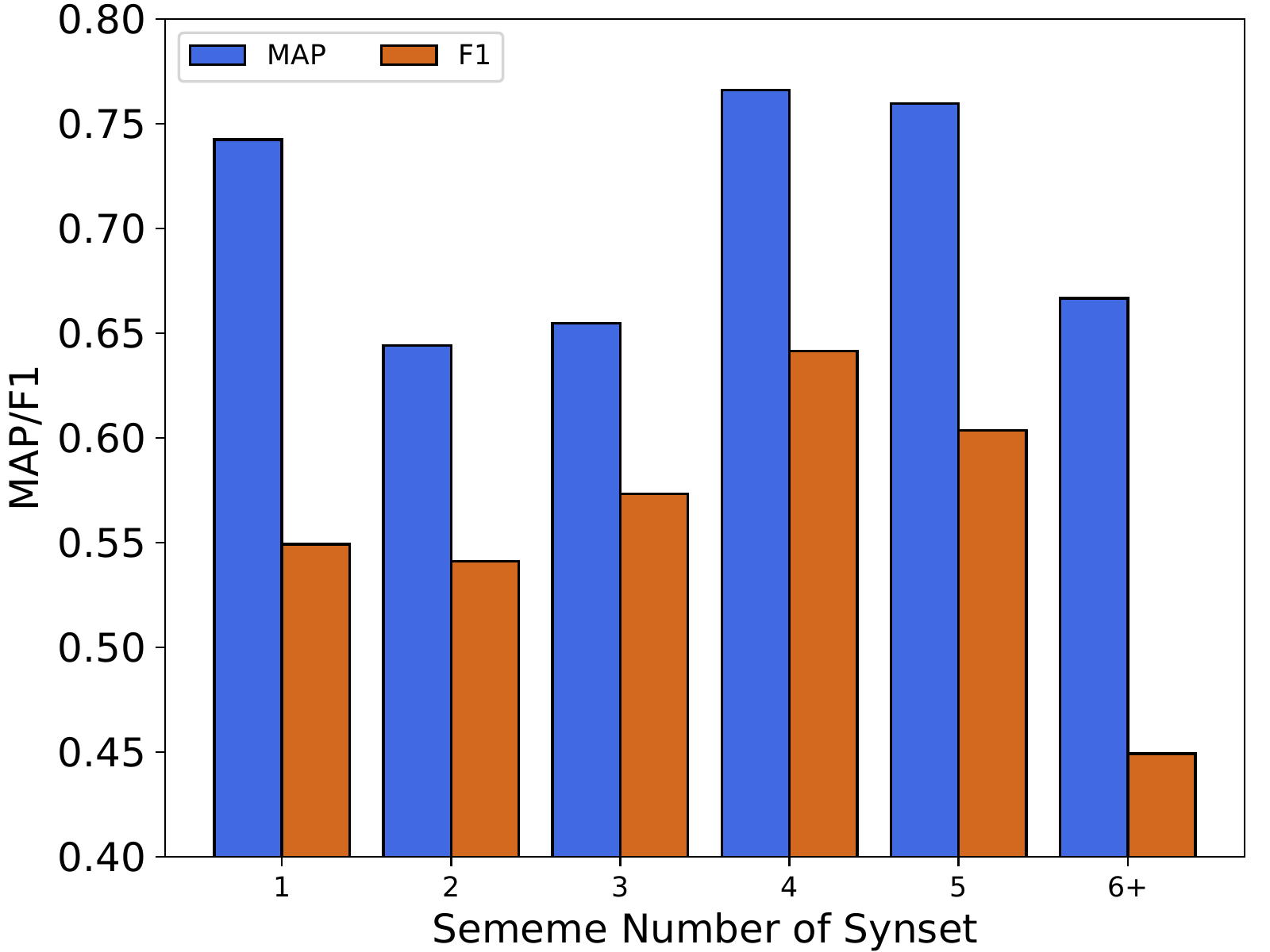}
    \caption{SPBS results of synsets whose sememe numbers are in different ranges. The numbers of synsets in the six ranges are 218, 239, 179, 179, 88 and 65 respectively.}
    \label{fig:eff-sememe-number}
\end{figure}

\subsection{Effect of the Synset's Sememe Number}
In this subsection, we explore whether the number of a synset's annotated sememes affects SPBS results. Figure \ref{fig:eff-sememe-number} exhibits the sememe prediction results of synsets whose sememe numbers are in different ranges. 
We find that sememe prediction performance increases first and then decreases with sememe numbers basically, which shows the synsets with too few or too many sememes are hard to cope with. 
The anomaly that MAP is high for the single-sememe synsets is because of the characteristic of MAP.\footnote{A single-sememe synset's sememe prediction MAP would be 1 as long as the correct sememe is ranked first, no matter how many sememes are selected as predicted results.}

\subsection{Effect of the Sememe's Degree}
\label{sec:sememedegree}
\begin{figure}[!t]
    \centering
    \includegraphics[width=.92\columnwidth]{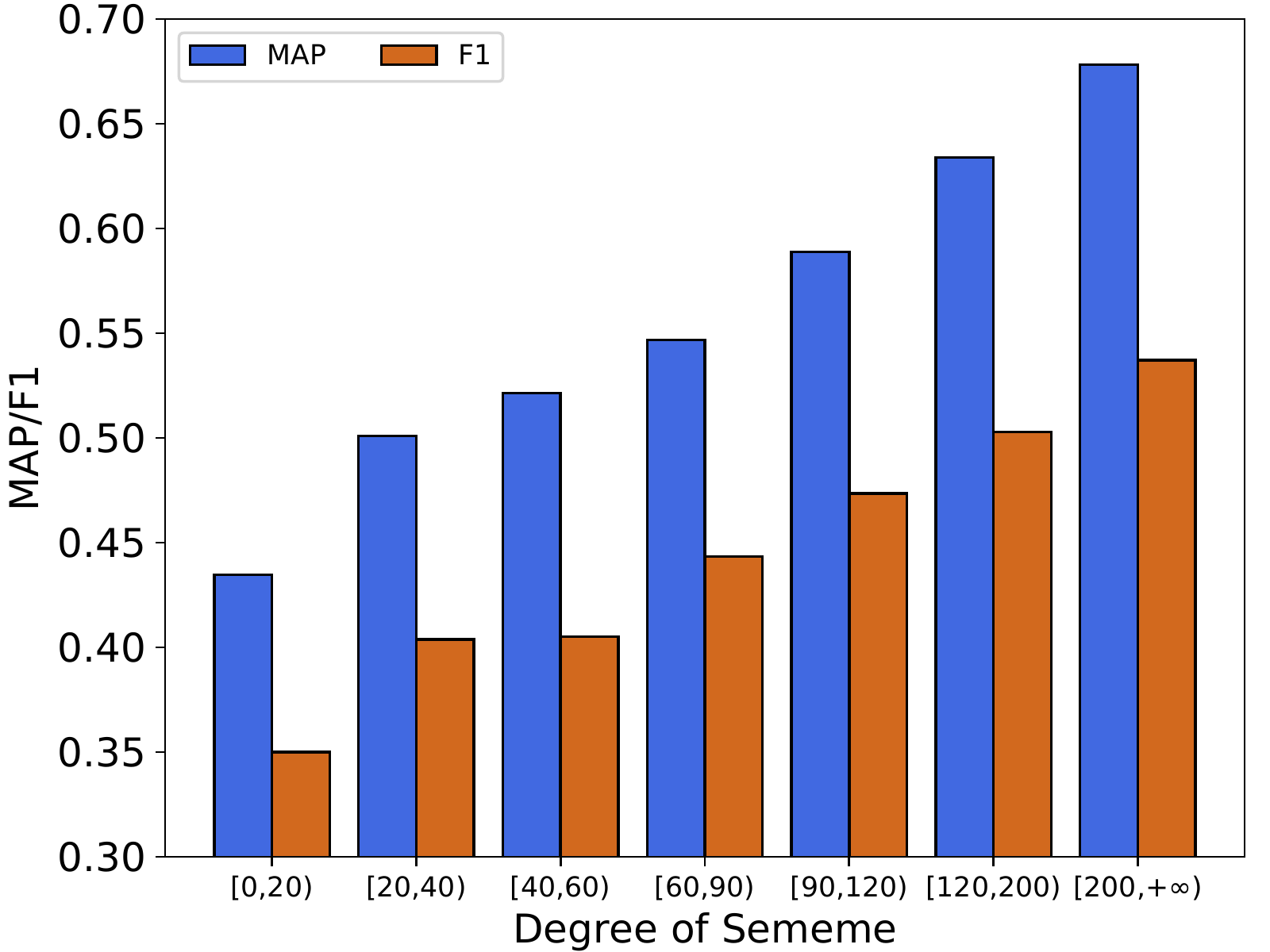}
    \caption{Average SPBS results of the synsets having sememes with degrees within different ranges. The numbers of sememes in the seven ranges are 1186, 235, 68, 47, 32, 26 and 28 respectively.}
    \label{fig:eff-sememe-degree}
\end{figure}

To investigate what sememes are easy or hard to predict, we first focus on the degree of a sememe.
Similar to the degree of a synset, the degree of a sememe is the number of the triplets comprising the sememe, and it is the only quantitative feature of sememes.
Figure \ref{fig:eff-sememe-degree} shows the experimental results, where sememe degree is on the x-axis and the average prediction performance of the synsets having sememes within corresponding degree ranges is on the y-axis. 
We can observe that it is harder to predict the sememes with lower degrees, whose reason is similar to that of low-degree synsets' bad performance, i.e., low-degree nodes in graph normally have poor representations.

\subsection{Effect of Sememe's Qualitative Features}


    
In this subsection, we intend to observe some typical sememes to find qualitative features of sememes influencing SPBS results.
Table \ref{tab:sememe-case} lists top 10 easiest and top 10 hardest sememes to predict as well as the average sememe prediction results of the synsets having them. 
We find most of the easiest sememes are concrete and normally annotated to the tangible entities.
For example, the easiest sememe \texttt{capital} is always annotated to the synsets of capital cities like ``Beijing''. 
As for the hardest sememes, they are more abstract, e.g., \texttt{expression} and \texttt{protect}, and usually annotated to the intangible concepts or non-nominal synsets.
Therefore, we speculate that the concreteness of sememes is also an important factor in SPBS results.

\section{Related Work}
\paragraph{HowNet}
As the most well-known sememe KB, HowNet has attracted considerable research attention. 
Most related work employs HowNet 
in specific NLP tasks \citep{liu2002,duan2007word,fu2013multi,niu2017improved,gu2018language,qi2019modeling,zang2019textual,qin2019enhancing,zhang2020multi}.
Some work tries to expand HowNet by predicting sememes for new words \citep{xie2017lexical,jin2018incorporating}. 
To the best of our knowledge, only \citet{Qi2018Cross} make an attempt to build a sememe KB for another language by cross-lingual lexical sememe prediction (CLSP). 
They learn bilingual word embeddings in a unified semantic space, and then predict sememes for target words according to their meaning-similar words in the sememe-annotated language.
However, CLSP can predict sememes for only one language at one time and 
cannot work on low-resource languages or polysemous words. 

\paragraph{BabelNet}
BabelNet \citep{navigli2012babelnet} is a multilingual encyclopedic dictionary which amalgamates WordNet \citep{miller1995wordnet} with Wikipedia as well as many other KBs, such as Wikidata \citep{vrandevcic2014wikidata} and FrameNet \citep{baker1998berkeley}. 
It has been successfully utilized in all kinds of tasks \citep{moro2013integrating,iacobacci2015sensembed,de2015semantics}, especially the cross-lingual or multilingual tasks \citep{navigli2012joining,vyas2016sparse}.
BabelNet has many advantages in terms of serving as the base of a multilingual sememe KB including 
(1) covering all the commonly used languages (284 languages); 
(2) incorporating polysemous words into multiple BabelNet synsets which enables sememe annotation for senses of polysemous words; 
and (3) amalgamating various resources, including dictionary definitions, semantic relations from WordNet and the content of Wikipedia pages, all of which can assist sememe annotation. 



\paragraph{Knowledge Graph Embedding}
There are innumerable methods of knowledge graph embedding (KGE) towards knowledge base completion or link prediction \citep{wang2017knowledge}. 
For example, translational distance models \citep{Bordes2013,Lin2015a,xiao2016transg}, semantic matching models \citep{nickel2011three,yang2014embedding,trouillon2016complex} and neural network models \citep{dettmers2018convolutional,Nguyen2018}.
However, none of these models consider heterogeneous knowledge graphs like the synset-sememe semantic graph. 
To our best knowledge, we are the first to model synsets and sememes in a semantic graph and propose a specifically modified KGE model for it.

\begin{table}[!t]
  \centering
  \resizebox{.95\columnwidth}{!}{
    \begin{tabular}{cc|cc}
    \toprule
    \multicolumn{2}{c|}{{10 Easiest Sememes}} & \multicolumn{2}{c}{{10 Hardest Sememes}} \\
    \midrule
    {Sememe} & {MAP/F1} & {Sememe} & {MAP/F1} \\
    \midrule
    {\texttt{capital}} & {96.4/81.3} & {\texttt{shape}} & {13.3/5.0} \\
    {\texttt{metal}} & {95.1/81.1} & {\texttt{document}} & {32.7/37.8} \\
    {\texttt{chemistry}} & {92.8/77.8} & {\texttt{expression}} & {36.5/30.1} \\
    {\texttt{city}} & {92.1/79.3} & {\texttt{artifact}} & {37.6/36.9} \\
    {\texttt{physics}} & {89.6/72.6} & {\texttt{protect}} & {37.9/31.6} \\
    {\texttt{provincial}} & {89.4/75.5} & {\texttt{animate}} & {38.6/27.8} \\
    {\texttt{PutOn}} & {87.8/56.9} & {\texttt{route}} & {40.4/32.3} \\
    {\texttt{place}} & {87.3/73.4} & {\texttt{implement}} & {43.8/45.1} \\
    {\texttt{ProperName}} & {86.6/71.5} & {\texttt{kind}} & {45.7/30.6} \\
    {\texttt{country}} & {85.7/63.3} & {\texttt{own}} & {47.3/56.0} \\
    \bottomrule
    \end{tabular}%
    }
  \caption{Top 10 sememes which are easiest to predict and top 10 sememes which are hardest to predict, and their corresponding sememe prediction results.}
  \label{tab:sememe-case}%
\end{table}%

\section{Conclusion and Future Work}
In this paper, we propose and formalize a novel task of Sememe Prediction for BabelNet Synsets (SPBS), which is aimed at facilitating the construction of a multilingual sememe KB based on BabelNet.
We also build a dataset named BabelSememe which serves as the seed of the multilingual sememe KB. 
In addition, we preliminarily propose two simple and effective SPBS models.
They utilize different kinds of information and can be combined to achieve better performance. 
Finally, we conduct quantitative and qualitative analyses, aiming to point out characteristics and difficulties of the SPBS task. 

In the future, we will try to use more useful information of BabelNet synsets, e.g., WordNet definitions, in the SPBS task to improve performance. 
We also consider predicting the hierarchical structures of sememes for BabelNet sysnets. 
Moreover, we will conduct extrinsic evaluations of the predicted sememes when there are enough annotated synsets.

\section{Acknowledgements}
This research is jointly supported by the Natural Science Foundation of China (NSFC) project under the grant No. 61661146007 and the NExT++ project, the National Research Foundation, Prime Minister’s Office, Singapore under its IRC@Singapore Funding Initiative. 
We also thank the anonymous reviewers for their valuable comments.

\fontsize{9.2pt}{10.2pt} \selectfont
\bibliographystyle{aaai} 
\bibliography{AAAI-QiF.9763} 


\end{document}